\documentclass[letterpaper, 10 pt, conference]{ieeeconf}  
\IEEEoverridecommandlockouts         
\usepackage{epsfig}
\usepackage{graphicx}
\usepackage{amsmath}
\usepackage{amssymb}
\usepackage{booktabs}
\usepackage{xcolor}
\usepackage{bbding}
\usepackage{colortbl}
\usepackage{algorithm}
\usepackage{algorithmic}
\usepackage{indentfirst}
\usepackage{multirow}
\usepackage{cite}
\usepackage{url}
\usepackage[colorlinks,linkcolor=blue]{hyperref}
\usepackage[utf8]{inputenc}
\usepackage[switch]{lineno}

\title{\LARGE \bf
MV-MOS: Multi-View Feature Fusion for 3D Moving Object Segmentation}

\author{
Jintao Cheng$^{\dagger}$, Xingming Chen$^{\dagger}$, Jinxin Liang, Xiaoyu Tang, Xieyuanli Chen and Dachuan Li$^*$ 
\thanks{J. Cheng and D. Li are with Research Institute of Trustworthy Autonomous Systems, and Department of Computer Science and Engineering, Southern University of Science and Technology, Shenzhen 518055, China.
}
\thanks{J. Cheng, X. Chen, J. Liang and X. Tang are with the School of Electronic and Information Engineering, South China Normal University, Foshan 528225, China. 
}
\thanks{X. Chen is with the College of Intelligence Science and Technology, National University of Defense Technology, Changsha, China. 
}
\thanks {$^{\dagger}$ J. Cheng and X. Chen contributed  equally  to  this  work.}
\thanks{$^{*}$ Corresponding author: Dachuan Li (lidc3@mail.sustech.edu.cn)}
\thanks{This work is supported in part by the National Natural Science Foundation of China under Grant 52272419 and in part by the Shenzhen Science and Technology Program (KJZD20230923114220042).}
}

\begin{document}

\maketitle
\thispagestyle{empty}
\pagestyle{empty}

\begin{abstract}
Effectively summarizing dense 3D point cloud data and extracting motion information of moving objects (moving object segmentation, MOS) is crucial to autonomous driving and robotics applications. 
How to effectively utilize motion and semantic features and avoid information loss during 3D-to-2D projection is still a key challenge. 
In this paper, we propose a novel multi-view MOS model (MV-MOS) by fusing motion-semantic features from different 2D representations of point clouds. To effectively exploit complementary information, the motion branches of the proposed model combines motion features from both bird's eye view (BEV) and range view (RV) representations. In addition, a semantic branch is introduced to provide supplementary semantic features of moving objects. Finally, a Mamba module is utilized to fuse the semantic features with motion features and provide effective 
guidance for the motion branches. 
We validated the effectiveness of the proposed multi-branch fusion MOS framework via comprehensive experiments, and our proposed model outperforms existing state-of-the-art models on the SemanticKITTI benchmark. The implementation codes are available at \href{https://github.com/Chengjt1999/MV-MOS}{https://github.com/Chengjt1999/MV-MOS}.
\end{abstract}

\section{Introduction}
 The accurate identification of surrounding moving objects is fundamental to autonomous driving and robotics applications as it directly affects the safety and reliability of the system. In addition, the properties and state information of moving objects are essential to many downstream tasks such as high-definition map building, scene understanding and decision-making. The 3D MOS task aims to distinguish moving objects from static entities by segmenting the perceived LiDAR point clouds. 
 
 Existing MOS models can be generally categorized into 3D voxel-based\cite{2024InsMOS, 20244DMOS, 2024Cylinder3D,2024spsequencenet} and 2D projection-based methods \cite{2024MotionSeg3D,2024MotionBEV,2024MFMOS,2024RVMOS}. To correctly classifying moving objects, they summarize and extract motion features from dense 3D point cloud data using different representations. The  Voxel-based methods typically voxelize dense point clouds into grids that are directly processed in 3D space. However, performing 3D convolution operations on high-dimensional data are computationally demanding, making it difficult to deploy in real-world applications where real-time performance is crucial. In contrast to computation intensive voxel-based approaches, 2D projection-based deep learning models have recently emerged as a promising pipeline for MOS. 
 They firstly map 3D point cloud data into a 2D representation through the BEV or range view projection. In this manner, 2D convolution operations can be effectively performed on the projected feature representations. 
\begin{figure}[!t]
	\centering
	\includegraphics[scale=0.41]{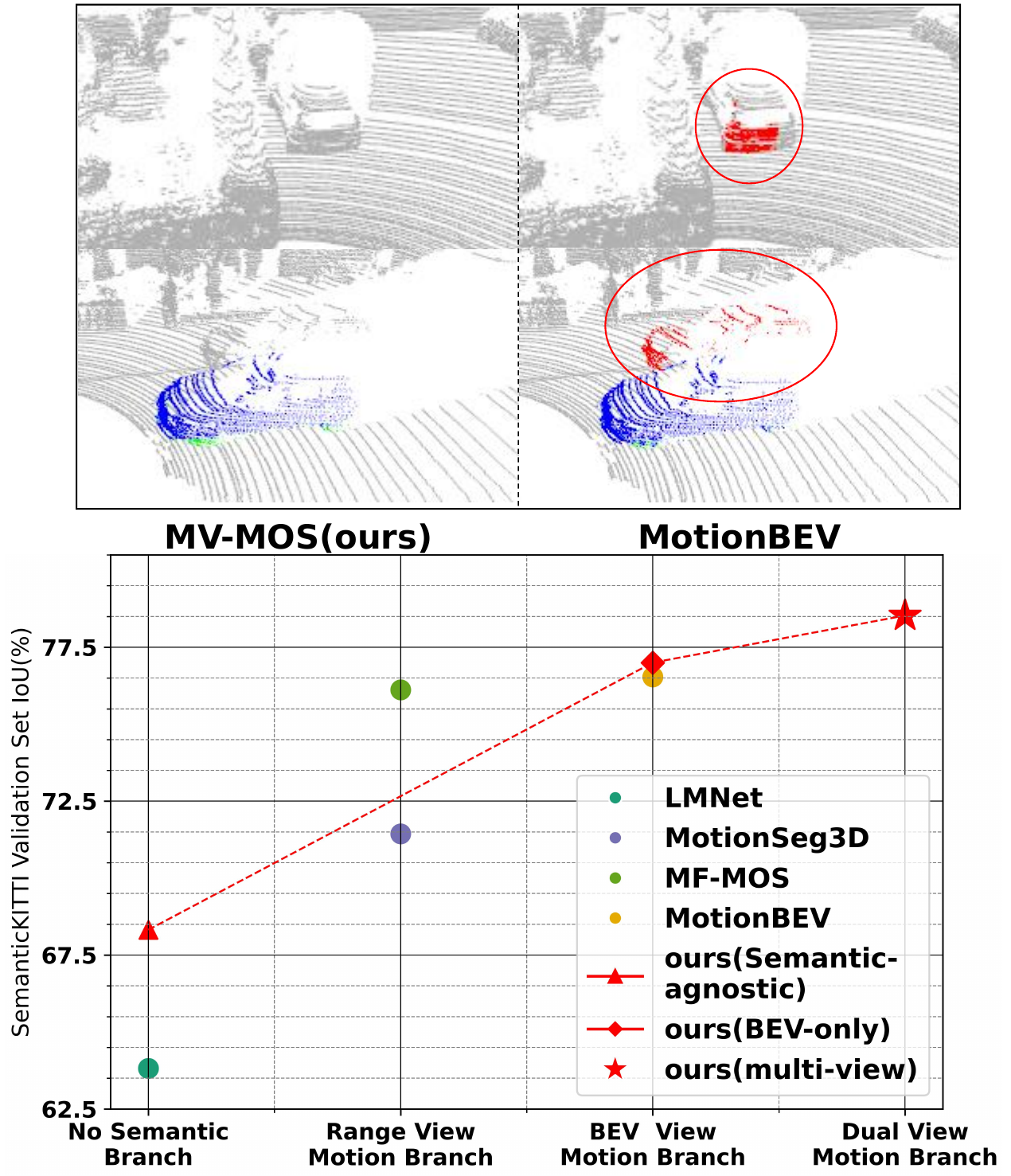}
	\caption{\textit{Upper}: Comparison of 3D moving object results from our proposed MV-MOS (\textit{upper left}) and the baseline MotionBEV \cite{2024MotionBEV} (\textit{upper right}), the segmented moving objects and incorrect segmentation (e.g. parked cars that are identified as moving cars) are colored in blue and red, respectively. \textit{Lower}: Comparison of performance of MOS models with different branch design on the SemanticKITTI-MOS benchmark, our proposed multi-view fusion model achieves the highest IoU.}
 \vspace{-14pt} 
\label{fig1}
\end{figure}
However, such 3D-to-2D transformation operation inevitably leads to information loss, affecting the quality of extracted feature and the ultimate segmentation accuracy.
For instance, when relying solely on the range view's residual map for motion information, stationary objects suddenly appearing in view due to occlusion by moving objects may be mistaken for moving objects. The BEV residual map relies only on height differences between frames for motion information, but the objects' own vertical information is ignored in the differencing operation, ultimately resulting in the loss of valuable motion contour information in the BEV motion feature map.
Therefore, how to effectively exploit the motion and semantic features of moving objects in different representation modalities is still a significant challenge to 3D MOS models. 

To address this challenge, we propose the Multi-view feature fusion MOS (MV-MOS), a novel 3D LiDAR MOS model, which features a dual-view and multi-branch structure for effective feature fusion. 
The proposed model can effectively retain rich and complementary motion and semantic information when using 2D  representations of 3D point clouds, while at the same time exploiting the well-developed 2D neural network models for lightweight and efficient feature processing. The primary contributions of this work are summarized as follows:

\begin{itemize}
\item We design a multi-branch structure that effectively exploits motion features of moving objects from both BEV and range view representations of point clouds.

\item We propose a novel feature synthesis neural network structure to comprehensively utilize complementary information from motions and appearance of moving object segmentation, and effectively generate rich semantic-guided motion features. In addition, we design an Mamba-based adaptive feature fusion framework which can robustly generate ultimate features for accurate segmentation, while addressing the issue of uneven feature density during fusion.

\item The proposed approach achieves 78.5\% and 80.6\% of IoU on the validation and test sets of the SemanticKITTI-MOS benchmark, respectively, surpassing state-of-the-art open-source MOS models.
\end{itemize}

\section{Related Work}
Existing MOS approaches can be categorized as non-learning-based and learning-based models. 
Conventional non-learning-based methods do not require complex data preprocessing and long training times. For example, \cite{2024peopleremover,2024robust} use the occupancy map method to calculate motion information based on volume occupancy differences. \cite{2024online} adopts the visibility-based theory and applies visual projection data to design a static points construction algorithm. Although these methods are relatively convenient, their segmentation accuracy is far inferior to that of deep learning-based methods.

For state-of-the-art learning-based methods, Cylinder3D\cite{2024Cylinder3D} employs cylindrical voxelization of point cloud data and uses sparse 3D convolution for feature processing. InsMOS\cite{2024InsMOS} and 4DMOS\cite{20244DMOS} introduce time series into the 3D space and use 4D convolution to enhance the processing of point cloud temporal information. LMNet\cite{2024LMNet} was the initially proposed model which introduces 2D projection in the range view to construct residual maps for capturing motion information. Similarly, models such as  MotionSeg3D\cite{2024MotionSeg3D} and MF-MOS\cite{2024MFMOS} are also built upon 2D representations, and they incorporate semantic branches and significantly improve segmentation accuracy. MotionBEV\cite{2024MotionBEV} adopts the BEV perspective for 2D projection, identifying moving objects based on point cloud height differences. However, such models are still subject to the high computational burden of 3D/4D convolutions and the inevitable information loss in single-view 2D representations. Motivated by such issues in MOS, we propose a multi-branch MOS neural network structure that combines BEV and range views. Our proposed structure leverages efficient 2D convolutional neural network models and the adaptive fusion module to effectively capture rich semantic and motion information from multiple representations.

\section{Methodology}
\subsection{Data Prepossessing}
The original LiDAR point cloud data $\boldsymbol{\boldsymbol(x,y,z)}$ are firstly transformed into range view and BEV presentations to derive the projected mappings $\boldsymbol{\boldsymbol{I}_{RV}^{i}(u, v)}$  and $\boldsymbol{\boldsymbol{I}_{BEV}^{i}(\widetilde{u}, \widetilde{v})}$, where $\boldsymbol{\boldsymbol(u,v)}$ and $\boldsymbol{\boldsymbol(\widetilde{u}, \widetilde{v})}$ represent coordinates in the 2D images, and 
$\boldsymbol{i}$ denotes a specific $i$th frame of point cloud data. $\boldsymbol{\boldsymbol{I}_{BEV}(\widetilde{u}, \widetilde{v})}$ will also be used as the input $\boldsymbol{\boldsymbol{X}_{Semantic}}$ for the semantic branch (the 3D-2D transformation process follows the standard setup of  methods in \cite{2024MotionSeg3D,2024MotionBEV}).

The range view and BEV residual maps for the $k$-th frame and the $0$-th frame are represented by the following formulas:
\begin{equation}
\boldsymbol{I}_{res-rv}^{k-0}(\boldsymbol{u}, \boldsymbol{v})=\left|\frac{\boldsymbol{I}_{R V}^k(\boldsymbol{u}, \boldsymbol{v})-\boldsymbol{I}_{R V}^0(\boldsymbol{u}, \boldsymbol{v})}{\boldsymbol{I}_{R V}^0(\boldsymbol{u}, \boldsymbol{v})}\right|
\label{I_rv}
\end{equation}
\begin{equation}
\boldsymbol{I}_{res-bev}^{k-0}(\widetilde{u}, \widetilde{v})=\left|{\boldsymbol{I}_{BEV}^k(\widetilde{u}, \widetilde{v})-\boldsymbol{I}_{BE V}^0(\widetilde{u}, \widetilde{v})}\right|
\label{I_bev}
\end{equation}
where  $\boldsymbol{\boldsymbol{I}_{res-bev}(\widetilde{u}, \widetilde{v})}$ and
$\boldsymbol{\boldsymbol{I}_{res-rv}(\boldsymbol{u}, \boldsymbol{v})}$ will also be used as the input $\boldsymbol{\boldsymbol{X}_{bev}}$ and $\boldsymbol{\boldsymbol{X}_{rv}}$ for the motion branch.

To improve robustness, a stacked frame representation $\boldsymbol{\boldsymbol{I}_{BEV}^{k^{\prime}}(\widetilde{u}, \widetilde{v})}$ is utilized in this work to replace the $\boldsymbol{\boldsymbol{I}_{BEV}^{k}(\widetilde{u}, \widetilde{v})}$ in the above equation: 
\begin{equation}
\boldsymbol{I}_{BEV}^{k^{\prime}}(\widetilde{u}, \widetilde{v})=\operatorname{Max}\{\boldsymbol{Z}(\widetilde{u}, \widetilde{v})\}-\operatorname{Min}\{\boldsymbol{Z}(\widetilde{u}, \widetilde{v})\}
\label{stk_bev}
\end{equation}
where $\boldsymbol{\boldsymbol{Z}(\widetilde{u}, \widetilde{v})}$  represents all elevation information within the  $\boldsymbol{k^{\prime}}$-frames window.

Since the projected views $\boldsymbol{\boldsymbol{I}_{RV}(u, v)}$ and $\boldsymbol{\boldsymbol{I}_{BEV}(\widetilde{u}, \widetilde{v})}$  of the two views both originate from the coordinates $\boldsymbol{(x,y,z)}$ in the 3D point cloud space, we can derive the transformation matrices (denoted as $\boldsymbol{Matrix_{r2b}}$ and $\boldsymbol{Matrix_{b2r}}$) based on this relationship. Converting the representation $\boldsymbol{\boldsymbol{I}_{RV}(u, v)}$ of the range view to the BEV view representation $\boldsymbol{\boldsymbol{I}_{r2b}(\widetilde{u}, \widetilde{v})}$  can be simply achieved by:
\begin{equation}
\boldsymbol{I}_{r2b}(\widetilde{u}, \widetilde{v})=\boldsymbol{F}_{\text {grid-sample}}(\boldsymbol{I}_{rv}(u, v),\boldsymbol{Matrix}_{r2b})
\label{r2v}
\end{equation}

The BEV and RV residual maps built using Eq. \ref{I_rv}-\ref{stk_bev} can be fed to the motion and semantic branches in the proposed model for further feature extraction. 
(Details of the derivation process can be found in \cite{2024GFNet}). 

\subsection{Network Structure}
\begin{figure*}[htbp]
    \vspace{+.2cm}
    \centering
    \includegraphics[width=0.95\linewidth]{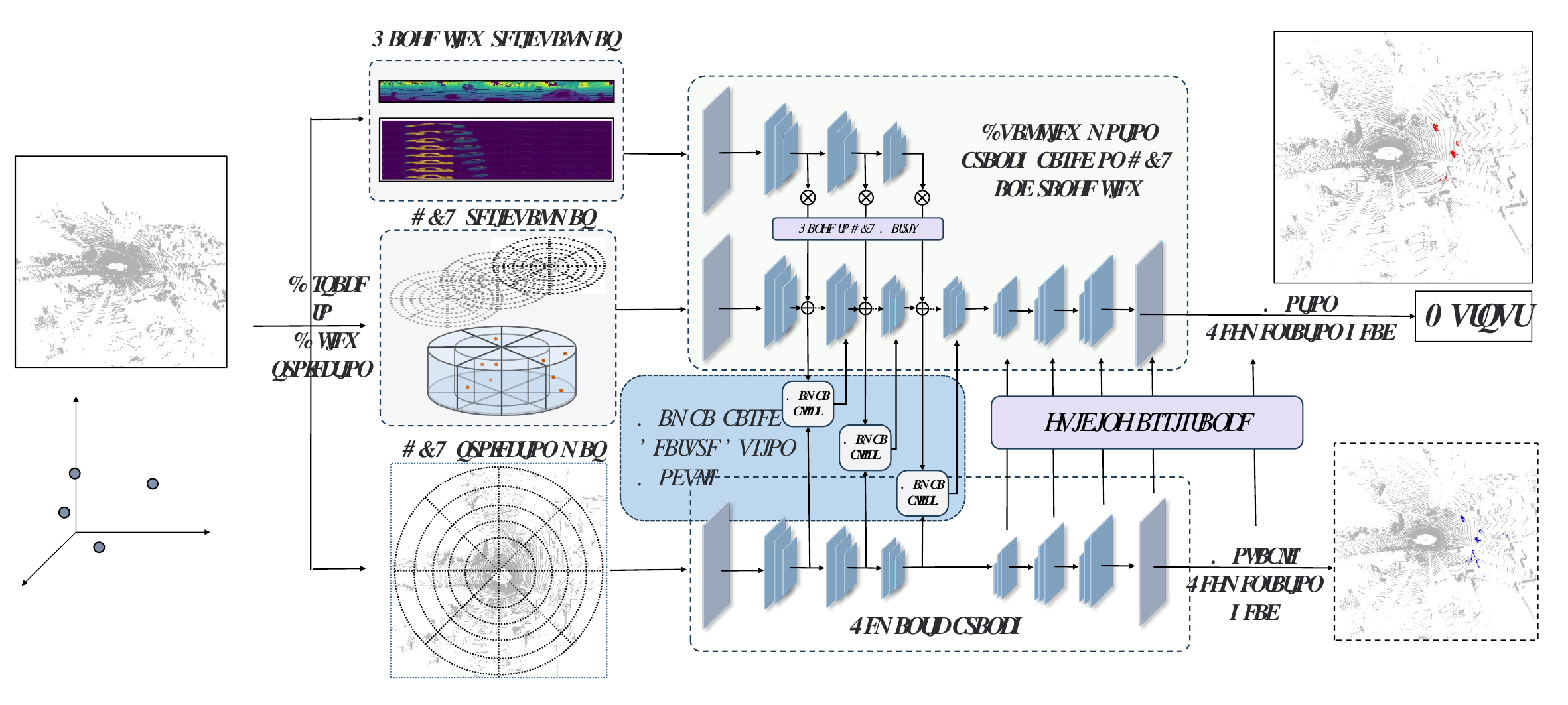}
   \caption{Overview of the proposed MV-MOS framework. In the motion branch, the motion information of moving objects are derived  by fusing two residual map features from the BEV and range view representations of LiDAR point clouds. The semantic branch extracts rich appearance features that supplement and guide the motion branch. The Mamba-based feature fusion module generates the synthesized features for the prediction of the final output.}
\label{overview}
\vspace{-.5cm}
\end{figure*}
\subsubsection{Motion Branch Structure Based on Muti-View Residual Map Fusion}
Due to the lightweight design of the UNet model, it can facilitate the additional overhead introduced by multi-branch networks. Therefore, we build our proposed model on the backbone of UNet, as shown in Fig.~\ref{overview}. The proposed motion-focused network backbone consists of two primary sub-branches for extracting motion features by constructing two residual maps from the BEV view and the range view. 
By analyzing the two 2D representations of the point cloud 3D coordinate system  $\boldsymbol{(x,y,z)}$, one can conclude that the motion information in the BEV view comes from the value difference of $\boldsymbol{z_i}$ at $\boldsymbol{(x_i,y_i)}$. In contrast, for the range view, the motion information can be derived from the depth value differences at the same azimuthal and polar angles, with the LiDAR as the origin. The residual maps of the two 2D view representations capture motion information from different perspectives, and such information from BEV and RV complement each other. 
Therefore, the designed motion branch of proposed MBR-MOS combines the two perspectives to achieve complementary information effectively.
For the construction of the input RV residual map $\boldsymbol{X}_{{rv}_{i}}$, cyclic convolutions are used sequentially for down-sampling, which is given by:
\begin{equation}
\boldsymbol{X}_{{rv}_{i+1}}=\boldsymbol{f}_{\text {circular}_{conv}}(\boldsymbol {Pool}({X}_{{rv}_{i}}))
\end{equation}
\begin{equation}
\boldsymbol{f}_{\text {circular}_{conv}}(x)=\boldsymbol{F}_{\text {RBC}}(\boldsymbol{Pad}(x)_\text {circular})
\end{equation}
where $\boldsymbol {Pool}$ represents down-sampling using a max pooling layer with a stride of 2. $\boldsymbol{Pad}(x)_\text {circular}$ indicates padding the feature map using a circular method. $\boldsymbol{F}_{\text {RBC}}$ denotes sequential processing through a 3x3 convolution operation, batch normalization layer, and ReLU activation function.

For the BEV residual map $\boldsymbol{X}_{{bev}_{i}}$  and the RV residual map $\boldsymbol{X}_{{rv}_{i}}$ of the same scale, the transformed view 
$\boldsymbol{X}_{{r2b}_{i}}$ is obtained using Eq. \ref{r2v}. We then fuse them and utilize a multi-channel attention mechanism to suppress invalid information, and keep the network focusing only on important representations. The process is given by:
\begin{equation}
	\begin{aligned}
\boldsymbol{X}_{{motion}_{i}}=\boldsymbol{X}_{{bev}_{i}} + \boldsymbol{Attention}_{hwc}(\\
\boldsymbol{f}_{\text {circular}_{conv}}(\boldsymbol{Cat}(\boldsymbol{X}_{{bev}_{i}},\boldsymbol{X}_{{r2b}_{i}})))
	\end{aligned}
\end{equation}
where $\boldsymbol{Cat}(\cdot)$ indicates concatenation of the feature maps along the channel dimension. $\boldsymbol{Attention}_{hwc}$ 
represents the attention computation performed on both the channel and spatial dimensions:
\begin{equation}
\boldsymbol{Attention}_{i \in(hw,c) }(x) = x \times \boldsymbol{F}_{\text {RBC}}(\boldsymbol{AvgPool}(x)_{i})
\end{equation}

The resulting motion feature map $\boldsymbol{X}_{{motion}_{i}}$ from the above process of the motion branch captures motion information from both perspectives. Such feature map is incorporated in the subsequent feature interaction and fusion with the semantic branch, which is described in the following subsection.

\subsubsection{Semantic Branch Structure Based on BEV Perspective Projection}
The proposed semantic branch serves two purposes in MV-MOS. Firstly, for the motion branch, the residual map contains information about the motion state of objects. Therefore, during the down-sampling stage of the backbone network, the point cloud projection feature map in the semantic branch adds rich object appearance semantic information to the motion branch, making the extraction of motion object features by the backbone network more accurate. 
This process is formulated as:
\begin{equation}
	\begin{aligned}
\boldsymbol{X}_{{fused}_{i}}=\boldsymbol{X}_{{motion}_{i}} + \boldsymbol{Attention}_{hwc}(\\
\boldsymbol{f}_{\text {circular}_{conv}}(\boldsymbol{Cat}(\boldsymbol{X}_{{motion}_{i}},\boldsymbol{X}_{{semantic}_{i}})))
	\end{aligned}
\end{equation}

Secondly, the semantic branch is responsible for predicting the movable attributes of objects and uses the same-sized semantic feature maps as guidance for the motion branch, enabling the primary motion branch to focus on detecting movable objects. This process involves two steps. First, the semantic branch performs up-sampling using the various sizes of $\boldsymbol{X}_{{semantic}_{i}}$:
\begin{equation}
\boldsymbol{X}_{{Sout}_{i}}=\boldsymbol{F}_{\text {up}}({X}_{{semantic}_{i+1}},{X}_{{semantic}_{i}})
\end{equation}
\begin{equation}
\boldsymbol{F}_{up}(X_{i+1},X_{i})=\boldsymbol{f}_{\text {circular}_{conv}}(\boldsymbol{Cat}(\boldsymbol{Up}(X_{i+1}),X_{i}))
\end{equation}
where $Up(\cdot)$ represents the PixelShuffle\cite{2024PixelShuffle} upsampling operation. The ultimate predicted output 
$\boldsymbol{X}_{Sout}$ is used  with the movable labels of objects to calculate the loss for training the motion branch, and enhance its capability of discerning object movability. In addition, the outputs $\boldsymbol{X}_{{Sout}_{i}}$ from the semantic branch upsampling layers are combined with the output feature map $\boldsymbol{X}_{{fused}_{i}}$ of the backbone network, providing guidance to the up-sampling stage of the motion branch. This process is formulated as:
\begin{equation}
	\begin{aligned}
\boldsymbol{X}_{{out}_{i}}=\boldsymbol{F}_{\text {up}}(\boldsymbol{X}_{{fused}_{i}},\boldsymbol{Cat}(\boldsymbol{X}_{Scout_{i}},\boldsymbol{X}_{fused_{i}}))
	\end{aligned}
\end{equation}

The final output $\boldsymbol{X}_{out}$ not only incorporates motion information from dual-view residual maps, but also integrates semantic information of objects separately in the down-sampling and up-sampling stages. Such design effectively enhances the model's capability to fully exploit features of moving objects.

\subsubsection{Density-aware Adaptive Feature Fusion Module}
\begin{figure}[!t]
	\centering
	\includegraphics[scale=0.24]{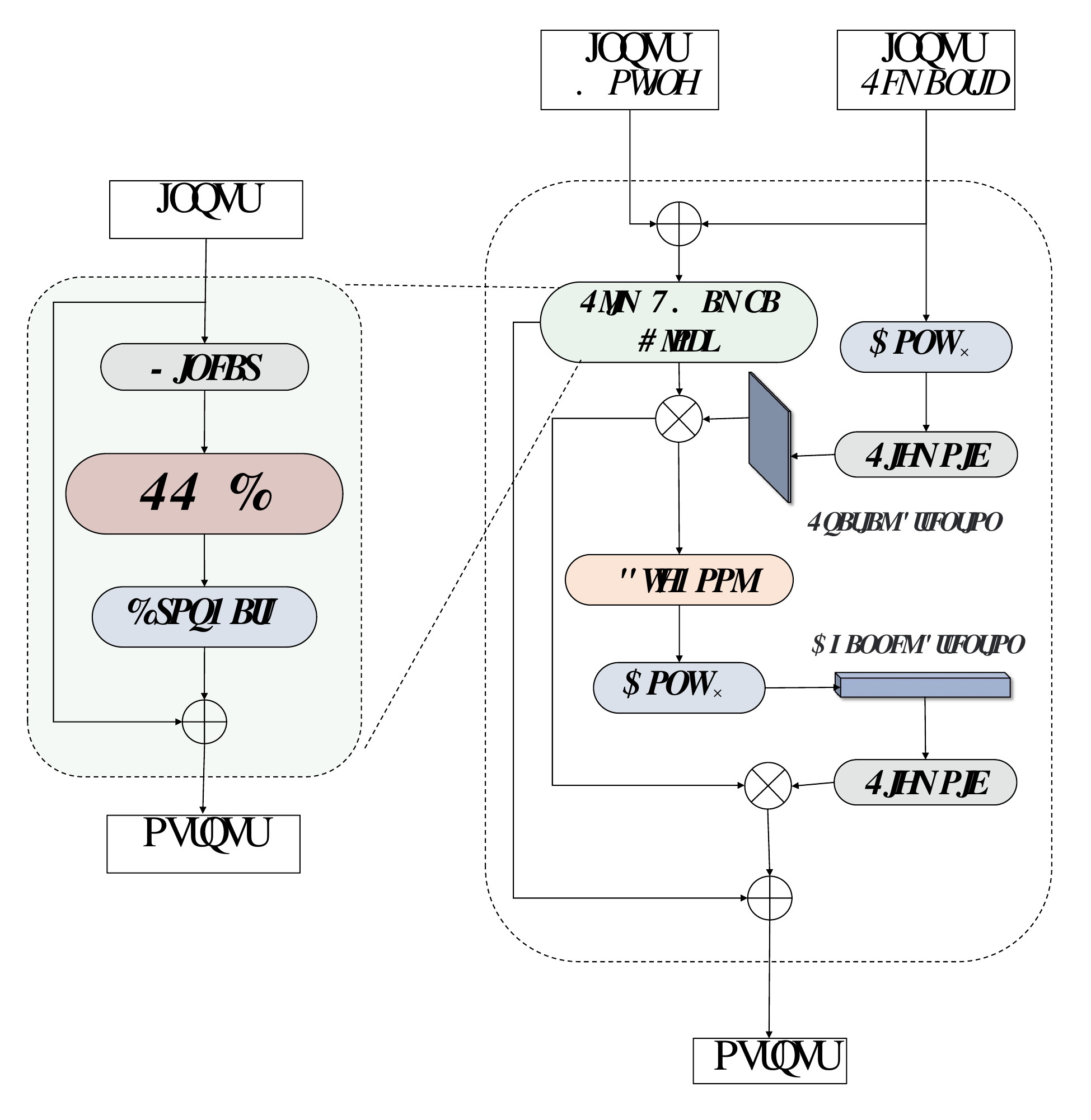}
     \vspace{-10pt} 
	\caption{Structure of the proposed adaptive feature fusion module.}
 \vspace{-14pt} 
\label{mamba}
\end{figure}
In the proposed dual-branch motion-semantic feature fusion structure, we utilize the motion stream as the primary branch and semantic information as the auxiliary branch. The proposed fusion module sequentially performs attention calculations on the spatial and channel dimensions of the semantic and the motion information input $\boldsymbol{F}_{s}$, $\boldsymbol{F}_{m}$, respectively. In this manner,  the semantic information can be used as an auxiliary to activate effective motion information in the motion feature map, while supplementing important contour information that might be lost in the residual map calculation. This process can be formulated as:
\begin{equation}
\boldsymbol{F}_{m}^{\prime}=\boldsymbol{F}_{m} \otimes  \boldsymbol{Sigmoid}(\boldsymbol{Conv}_{1 \times 1}(\boldsymbol{F}_{s}))
\end{equation}
\begin{equation}
	\begin{aligned}
\boldsymbol{F}_{out}=\boldsymbol{F}_{m} + \boldsymbol{F}_{m}^{\prime} \otimes  \boldsymbol{Softmax}(\\
\boldsymbol{Conv}_{1 \times 1}( \boldsymbol{AvgPool}(\boldsymbol{F}_{m}^{\prime})))
	\end{aligned}
\end{equation}

Compared with the semantic feature maps, the motion feature maps obtained from residual maps are sparser. As a result, directly fusing these two types of feature maps would cause the dense semantic features to dominate the synthesized feature, which may deteriorate the resulting MOS performance, as the semantic features are supposed to be auxiliary. 
To mitigate such feature imbalance, we designed an adaptive feature fusion mechanism based on the mamba structure \cite{2024Mamba}, which 
can dynamically adjust its state based on the model's current input and selectively retain important information in the sequence. 
Considering that the current information flow is represented as 2D images, the SS2D (consisting of Scan Expansion, S6 Block, and Scan Merge) mechanism proposed by Vmamba\cite{2024VMamba} is more suitable for the application in this work. 
Therefore, we introduce a Mamba mechanism based on SS2D and embed it into the semantic-motion dual-branch fusion module, as illustrated in Fig.~\ref{mamba}:
\begin{equation}
	\begin{aligned}
\boldsymbol{F}_{fused}=\boldsymbol{F}_{sm} + \boldsymbol{DropPath}(\\
\boldsymbol{SS2D}(\boldsymbol{Linear}(\boldsymbol{Cat}(\boldsymbol{F}_{sm}))))
	\end{aligned}
\end{equation}
where $\boldsymbol{F}_{sm}$ is derived from the concatenation of $\boldsymbol{F}_{s}$ and $\boldsymbol{F}_{m}$ , and $\boldsymbol{F}_{fused}$ will replace $\boldsymbol{f}_{m}$  in Eq. (13) and (14). In this manner, the proposed feature fusion process can focus on key elements and ignore uninformative redundant features.  

\subsection{Loss Function}
During the training process, we introduce two sets of labels: \textit{moving} objects and \textit{movable} objects, and the motion branch and semantic branches are trained separately. The total loss  $\boldsymbol{L}_{\text {Total}}$  is given by:

\begin{equation}
\boldsymbol{L}_{\text {Total}}=\boldsymbol{L}_{\text {Moving}}+\boldsymbol{L}_{\text {Movable}},
\end{equation}
where $\boldsymbol{L}_{\text {Moving}}$ and $\boldsymbol{L}_{\text {Movable}}$ represent the segmentation loss functions for moving objects and movable objects, respectively. Both $\boldsymbol{L}_{\text {Moving}}$ and $\boldsymbol{L}_{\text {Movable}}$ are composed of the cross-entropy loss function($\boldsymbol{L}_{\text {CE}}$) and the Lovász-Softmax function($\boldsymbol{L}_{\text {LS}}$), combined with the relevant labels for calculation:
\begin{equation}
\boldsymbol{Loss}=\boldsymbol{L}_{\text {CE}}+\boldsymbol{L}_{\text {LS}}
\end{equation}
\begin{equation}
\boldsymbol{L}_{\text {CE}} =-\sum_i^N y_i \cdot \log \left(p_i\right)
\end{equation}
where $\boldsymbol{N}$ is the number of samples, $\boldsymbol{y}_{i}$ is the ground truth label of the $\boldsymbol{i}$th sample (0 or 1), $\boldsymbol{p}_{i}$ is the predicted probability of the 
$\boldsymbol{i}$th sample being in the positive class. The Lovász-Softmax loss is defined by: 
\begin{equation}
\boldsymbol{L_{\text {LS}}}=\frac{1}{|C|} \sum_{c=1}^{|C|} \operatorname{Lov\acute{\text{a}}sz}\left(\Delta J_c\right)
\end{equation}
\begin{equation}
\operatorname{Lov\acute{\text{a}}sz}(\Delta J)=\sum_{i=1}^N \Delta J_{(i)}\left(m_{(i)}-m_{(i-1)}\right)
\end{equation}
where $\boldsymbol{|C|}$ is the number of classes in a multiple classification problem. $\boldsymbol{\Delta J_c}$  is the Jaccard loss subset associated with class $\boldsymbol{c}$, representing the change in IoU loss for each class. $\boldsymbol{\Delta J_{(i)}}$ is the IoU loss change for the $\boldsymbol{i}$-th sample after being sorted in descending order of the predicted error probability. $\boldsymbol{m_{(i)}}$ is the cumulative error for the $\boldsymbol{i}$-th sample after sorting.

\section{Experiments}
\subsection{Experiments Setups}
The proposed MV-MOS is trained and evaluated on the SemanticKITTI\cite{2024SemanticKITTIdataset}, which is a public LiDAR point cloud dataset for the MOS task. The dataset includes semantic labels for 28 classes of objects in real-world driving scenarios, such as pedestrians, cars, as well as labels for dynamic/static and movable attributes. 

Our proposed MV-MOS is implemented based on PyTorch 1.12.0 and all experiments are conducted on NVIDIA RTX 4090 and Tesla V100 GPUs. The number of training epochs is set to 100, with an initial learning rate of 0.01 and a decay factor of 0.95 per epoch. The batch size is set to 4 per GPU. During training, we use the SGD optimizer with a momentum of 0.9 and a weight decay of 0.0001. Following the standard evaluation practice in MOS tasks, we use the Intersection over Union (IoU) metric to quantify the performance of our proposed approach in all experiments.

\subsection{Evaluation Results and Comparisons}

\begin{table}[t]
\vspace{+.3cm}
\caption{Performance Comparison of Different MOS Models on SemanticKITTI-MOS dataset.}
\centering
\label{tab1}
{\begin{tabular}{lccc}
    \toprule  
    Methods & Publication & Val(\%)$\uparrow$  & Test (\%)$\uparrow$\\
    \midrule
    SpSequenceNet \cite{2024spsequencenet} & CVPR 2020 & - & 43.2 \\
    LMNet \cite{2024LMNet} & ICRA 2021  & 63.8 & 60.5\\
    Cylinder3D \cite{2024Cylinder3D} & CVPR 2021  & 66.3 & 61.2 \\
    4DMOS \cite{20244DMOS} & RAL 2022  & 71.9 & 65.2 \\
    MotionSeg3D \cite{2024MotionSeg3D} & IROS 2022 &  71.4  & 70.2\\
    RVMOS \cite{2024RVMOS} & RAL 2022 & 71.2 & 74.7\\
    InsMOS \cite{2024InsMOS} & IROS 2023 & 73.2 & 75.6 \\
    MotionBEV \cite{2024MotionBEV} & RAL 2023 & 76.5 & 75.8 \\
    MF-MOS \cite{2024MFMOS} & ICRA 2024 & 76.1 & 76.7 \\
    \midrule
    MV-MOS(ours) & - & \textbf{78.5} &  \textbf{80.6}\\
\bottomrule
\end{tabular}}
\end{table}

\begin{table}[t]
\vspace{+.3cm}
\caption{Ablation Experiment of Different Combinations of Components on SemanticKITTI-MOS validation Set.}
\centering
\label{tab3}
{\begin{tabular}{ccccc}
    \toprule
    Methods & \multicolumn{3}{c}{Component} & IoU (\%)$\uparrow$ \\
    \midrule
        & Se-Mo & BR-Mo & Mamba &  \\
        & -Branch & -Branch & Block &  \\
    LMNet \cite{2024LMNet} &- & - & - &63.82 \\
    MV-MOS (\textit{i}) & - & - & - & 68.32  \\
    MotionBEV \cite{2024MotionBEV}& \Checkmark* & - & - & 76.54  \\
    MV-MOS (\textit{ii}) & \Checkmark & - & - & 77.00  \\
    \midrule
    \multirow{3}{*}
    {MV-MOS (\textit{iii})} 
    &  \Checkmark * & \Checkmark & - & 77.38  \\
    & \Checkmark & \Checkmark & - & 77.66 \\
    & \Checkmark & - & \Checkmark & 78.14  \\
    \midrule
    MV-MOS &  \Checkmark & \Checkmark & \Checkmark & \textbf{78.53}  \\
\bottomrule
\end{tabular}}
\end{table}

\begin{table}[t]
\centering
\caption{Model inference time (ms) results.}
\label{tab5}
\begin{tabular}{ccccc}
\toprule 
InsMOS & MotionSeg3D & MF-MOS & MotionBEV & \textbf{MV-MOS}   \\
\toprule 
193.68 & 117.01 & 96.19 & 34.41 & 77.27 \\
\bottomrule
\end{tabular}
\label{comparation}
\vspace{-.5cm}
\end{table}

Table \ref{tab1} presents the comparison results between MV-MOS and other state-of-the-art methods in the MOS task. 
From the table, it can be seen that the baseline LMNet, which only uses a single motion branch, achieves an accuracy of only 63.8\% on the validation set. In contrast, recent SOTA models MotionBEV and MF-MOS, which combine motion and semantic branches, achieve IoU values of over 76\%, surpassing voxel-based methods. This demonstrates the significant potential of models based on 2D projection methods. Our proposed MV-MOS integrates multiple perspectives and branches to enhance the capability of 2D projection methods in capturing motion information from 3D point cloud data. The proposed MV-MOS achieves the highest accuracy of 78.5\% on the validation set
which outperforms all other baseline MOS models. 
In addition, we further tested the proposed model on the  official SemanticKITTI-MOS benchmark server, and our proposed MV-MOS currently currently ranks first among all open-source models, with an IoU of 80.6\%.

\begin{figure*}[htbp]
    \vspace{+.2cm}
    \centering
    \includegraphics[width=0.95\linewidth]{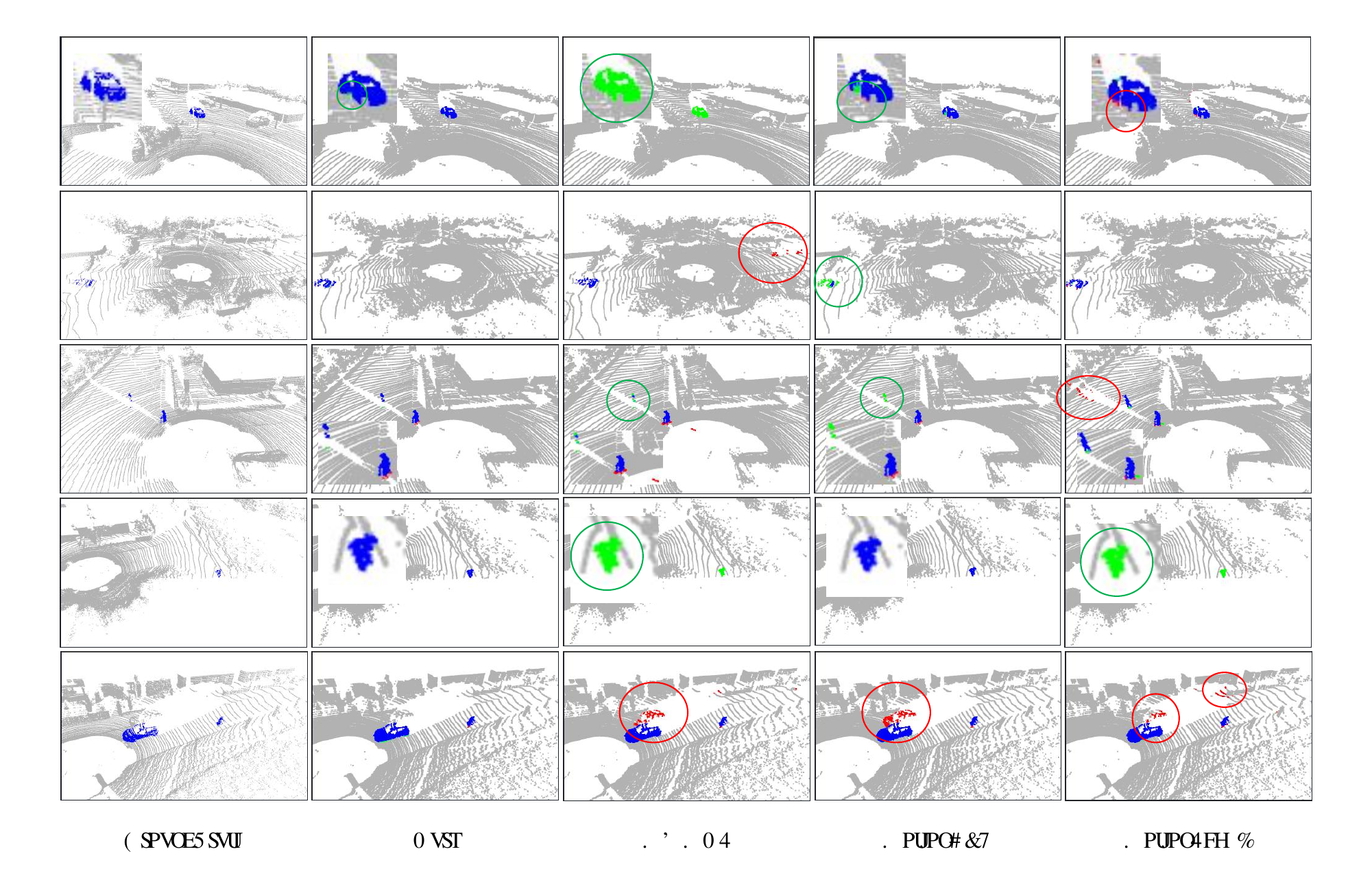}
   \caption{Qualitative moving object segmentation results of different models on the SemanticKITTI validation set. True positive, false negative, and false positive segmented points are colored in blue, green, and red, respectively. Incorrect segmentation results and missed moving objects are also highlighted in red and green circles, respectively (better view with color and magnification).}
\label{compare}
\vspace{-.5cm}
\end{figure*}

\subsection{Ablation Studies}
 We conducted further ablation experiments to investigate how the carefully designed individual structures and modules affect the performance of MV-MOS, with results shown in Table \ref{tab3}. 

The first set of experiments evaluate the benefit of the semantic information. We built two more mutated MOS models with different combination of feature extraction branches: the MV-MOS(i) represents the model without the semantic branch, with the motion branch based only on BEV residual maps; MV-MOS(ii) represents the introduction of only the motion-semantic dual-branch structure based on the BEV representations, with an additional guiding branch for movable attribute inference as an auxiliary branch. For a more intuitive comparison, we also introduced LMNet and MotionBEV as baseline approaches. As can be seen from the results (Table \ref{tab3}), the segmentation accuracy of MV-MOS(ii) and MotionBEV, which combine the semantic and motion branches, is about 10\% higher than that of the semantic-only LMNet and MV-MOS(i). 
Additionally, compared with the basedline MotionBEV, the motion-semantic multi-branch structure design in the proposed MV-MOS provides auxiliary feature-guidance, which result in an increase of IoU of nearly 0.5\%.

The second set of experiments aims to demonstrate the effectiveness of the combination of the RV and BEV view residual branches in the motion branch of MV-MOS (subsequently referred to as B(EV)R(ange View)-Motion-Branch), as well as the role of the adaptive fusion mechanism. 
As shown in Table \ref{tab3}, the first row of MV-MOS(iii) indicates that introducing the BR-Motion-Branch into a MotionBEV-like motion-semantic dual-branch structure improves the accuracy from 76.54\% to 77.38\%. Introducing the BR-Motion-Branch into MV-MOS(ii) increases the accuracy to 77.66\%. 
Additionally, incorporating our proposed Mamba Block at the connection point between the motion and semantic branches increases the accuracy to 78.14\%, an 1.14\% improvement over MV-MOS(ii). 

The third set of experiments demonstrates that our final model, MV-MOS, combining all modules, achieves the highest accuracy of 78.53\%. These three sets of experiments successively verify the effectiveness of the proposed modules from multiple aspects.

\subsection{Qualitative Analysis}
To more intuitively demonstrate the effectiveness of MV-MOS, we performed a qualitative analysis through visualization. Fig.~\ref{compare} compare the performance results of MV-MOS and several state-of-the-art models on the SemanticKITTI validation set. As shown, our model correctly infers more points, and the segmentation of objects is more complete (indicated by blue points). Additionally, when moving objects are at the edge of the field of view—either suddenly appearing or about to disappear—other models are prone to missed detections (green points) and false detections (red points). In contrast, our model, based on the multi-branch design, leverages motion information represented by multi-view 2D projections, which allows it to handle such situations more effectively. 

\subsection{Computational Efficiency}
All runtime testing experiments were conducted on a single Tesla V100 GPU for inference. As shown in Table \ref{tab5}, we compared MV-MOS with four state-of-the-art models from the past two years. In terms of inference time, MV-MOS significantly outperforms InsMOS\cite{2024InsMOS} and MotionSeg3D\cite{2024MotionSeg3D} and shows nearly a 20\% improvement over MF-MOS\cite{2024MFMOS}. Among all five compared models, MotionBEV\cite{2024MotionBEV} achieves the fastest per-frame processing speed at 34.41ms, surpassing all other models. The additional inference time of MV-MOS compared to MotionBEV\cite{2024MotionBEV} is anticipated due to the introduction of an extra range view motion branch and a guiding branch for inferring the movable attributes of objects. Given that MV-MOS consistently demonstrates significantly higher accuracy across different datasets compared to MotionBEV\cite{2024MotionBEV}, the added computational overhead is justifiable.
Considering that LiDAR typically operates at a frame rate of 10 Hz, and MV-MOS's per-frame inference speed is less than 100ms, it meets the requirement for real-time processing. Combined with its superior segmentation accuracy, this evidence highlights the superiority and practicality of our method in the moving object segmentation task.

\section{Conclusion}
In this paper, we propose a 
 novel 3D moving object segmentation neural network model based on multi-view and motion-semantic branch feature fusion. The proposed multi-branch BEV and range view fusion structure can effectively exploit complementary motion information of moving objects from different perspectives,  avoiding the loss of valuable information during the transformation of 3D point clouds to 2D representations. In addition, the proposed semantic branch and Mamba-based feature fusion design can effectively provide guidance and enhancement to the motion features, which ultimately improves the quality of synthesized features. Comprehensive comparative and ablation experiments validate the effectiveness and generalization of our proposed model. In addition, the proposed model also achieves desirable inference computation efficiency. Therefore, our proposed MV-MOS shows great potential for applications in practical autonomous driving and robotics systems. 


\bibliographystyle{IEEEtran}  
\bibliography{IEEEabrv,MSBR-MOS}
%
%

\end{document}